\documentclass[oribibl]{llncs}
\usepackage{multicol}
\usepackage{amsmath}
\usepackage{amssymb}
\usepackage{graphicx}
\usepackage{color}
\usepackage{hhline}
\usepackage{enumfrom}

\usepackage{hyperref}

\newcommand{\ignore}[1]{}

\newcommand{\mbf}[1]{\mathbf{ #1}}
\newcommand{\e}{\mathbf{e}}
\newcommand{\msfs}[1]{\mbox{\scriptsize{\sf  #1}}}
\newcommand{\msf}[1]{\mbox{\sf  #1}}

\newcommand{\red}[1]{\textcolor{red}{#1}}

\newcommand{\ul}[1]{\underline{ #1}}

\newcommand{\boxtheorem}{  \hfill $\Box$}
\newcommand{\nit}[1]{{\it #1}}

\newcommand{\comlb}[1]{{\vspace{2mm}\noindent \bf \red{COMM(LEO):}}~ #1   \hfill {\bf
    END.}\\}
\newcommand{\comgr}[1]{{\vspace{2mm}\noindent \bf COMM(GABBY):}~ #1   \hfill {\bf
    END.}\\}
    
\newcommand{\mc}[1]{\mathcal{ #1}}

\newcommand{\shap}{\mbox{\sf Shap}}

\newcommand{\es}{\mathbf{e}}
\newcommand{\resp}{\mbox{\sf Resp}}
\newcommand{\xresp}{\mbox{\sf x-Resp}}

\title{Answer-Set Programs for Reasoning about Counterfactual Interventions and Responsibility Scores for Classification }

\author{{\bf Leopoldo Bertossi}$^{1,2}$ and {\bf Gabriela Reyes}$^1$}
\institute{\bf Universidad Adolfo Ib\'a\~nez$^1$\\{\bf Faculty of Engineering and Sciences}\\ and\\ Millennium Inst. for Foundational Research on Data (IMFD)$^2$\\
 Santiago, \ Chile\\ leopoldo.bertossi@uai.cl \ and \ gabreyes@alumnos.uai.cl}

\ignore{
\author{{\bf Leopoldo Bertossi}}
\institute{\bf Universidad Adolfo Ib\'a\~nez\\{\bf Faculty of Engineering and Sciences}\\ and\\ Millennium Inst. for Foundational Research on Data (IMFD)\\
 Santiago, \ Chile\\ leopoldo.bertossi@uai.cl}
 }

\begin{document}

\thispagestyle{empty}
\pagestyle{plain}
\maketitle

\begin{abstract} \vspace{-4mm}
We describe how {\em answer-set programs} can be used to declaratively specify counterfactual interventions on entities under classification, and reason about them. In particular, they can be used to define and compute responsibility scores as attribution-based   explanations for outcomes from classification models. The approach allows for the inclusion of domain knowledge and supports query answering. A detailed example with a  naive-Bayes  classifier is presented.\vspace{-2mm}
\end{abstract}

\section{Introduction}

\vspace{-2mm}Counterfactuals are at the very basis of the notion of {\em actual causality} \cite{HP05}. They are hypothetical interventions (or changes) on variables that are part of a causal structural model. Counterfactuals can be used to define and assign {\em responsibility scores} to the variables in the model, with the purpose of quantifying their causal contribution strength to a particular outcome \cite{CH04,halpern15}. These generals notions of actual causality have been successfully applied in databases, to investigate actual causes and responsibilities for query results \cite{suciu,suciuDEBull,tocs}.

 Numerical scores have been applied in {\em explainable AI}, and most prominently with machine learning models for classification \cite{molnar}. Usually, feature values in entities under classification are given numerical scores, to indicate how relevant  those values are for the outcome of the classification. For example, one might want to know how important is the city or the neighborhood where a client lives when a bank uses a classification algorithm to accept or reject his/her loan request. We could, for example, obtain a large responsibility score for the feature value ``Bronx in New York City". As such it is a {\em local explanation}, for the entity at hand, and in relation to its participating feature values.

 A widely used score is $\msf{Shap}$ \cite{LetA20}, that is based on the Shapley value of coalition game theory \cite{R88}. As such, it is based on {\em implicit} counterfactuals and a numerical aggregation of the outcomes from classification for those different counterfactual versions of the initial entity. Accordingly, the emphasis is not on the possible counterfactuals, but on the final numerical score. However, counterfactuals are interesting {\em per se}. For example, we might want to know if the client, by changing his/her address, might turn a rejection into the acceptance of the loan request. The so generated new entity, with a new address and a new label, is a {\em counterfactual version} of the original entity.

 In \cite{tplp} the $\xresp$ score was introduced. It is defined in terms of explicit counterfactuals and responsibility as found in general actual causality. A more general version of it, the $\resp$ score, was introduced in \cite{deem}, and was compared with other scores, among them, $\msf{Shap}$. For simplicity we will concentrate on $\xresp$.

  Following up our interest in counterfactuals, we propose {\em counterfactual intervention programs} (CIPs). They are   {\em answer-set programs} (ASPs) \cite{asp,gelfond} that are used to specify counterfactual versions of an initial entity, and compute the $\xresp$ scores for its feature values. \ More specifically, here we present approaches to- and results about the use of ASPs   for specifying  counterfactual interventions on entities under classification, and reasoning about them. In this work, we show CIPs and their use in the light of a naive-Bayes classifier. \ See \cite{tplp} for more details and an example with a decision-tree classifier; and  \cite{rw21} for more examples of the use of ASPs for actual causality and responsibility.

 ASP is a flexible and powerful logic programming paradigm that, as such, allows for declarative specifications and reasoning from them.  The (non-monotonic) semantics of a program is given in terms of its {\em stable models}, i.e. special models that make the program true \cite{GL91}. In our applications, the relevant counterfactual versions correspond to different models of the CIP.
 \ In our example with a naive-Bayes classifier,   we use the {\em DLV} system \cite{leone} and its {\em DLV-Complex} extension \cite{calimeri08,calimeri09} that implement the ASP semantics; the latter with set- and numerical aggregations.

 CIPs can be used to specify the relevant counterfactuals, analyze different versions of them, and use them to specify and compute the $\xresp$ score. By using additional features of ASP, and of {\em DLV} in particular, for example {\em strong and weak program constraints}, one can specify and compute maximum-responsibility counterfactuals. The classifier can be specified directly in the CIP, or can be invoked as an external predicate \cite{tplp}. The latter case could be that of a {\em black-box classifier} \cite{rudin}, to  which $\msf{Shap}$ and $\xresp$ can be applied.

 CIPs are very flexible in that one can easily add {\em domain knowledge} or {\em domain semantics}, in such a way that certain counterfactuals are not considered, or others are privileged. In particular, one can specify {\em actionable counterfactuals}, that, in certain applications, make more sense and may lead to feasible changes of feature values for an entity to reverse a classification result \cite{ustun,karimi}. All these changes are much more difficult to implement if we use a purely procedural approach. With CIPs, many changes of potential interest can be easily and seamlessly  tried out on-the-fly, for exploration purposes.

 Reasoning is enabled by query answering, for which two semantics are offered. Under the {\em brave semantics} one obtains as query answers those that hold in {\em some} model of the CIP. This can be useful to detect if there is ``minimally changed" counterfactual version of the initial entity where the city is changed together with the salary. Under the {\em cautious semantics} one obtains answers that hold in all the models of the CIP, which could be used to identify feature values that do not change no matter what when we reverse the outcome. Query answering on ASPs offers many opportunities.

This paper is structured as follows. In Section \ref{sec:cla}, we introduce and discuss the problem, and provide an example. In Section \ref{sec:NB} we introduce the naive-Bayes classifier we will use as a running example. In Section \ref{sec:xresp} we define the $\xresp$ score. In Section \ref{sec:cips} we introduce {\em counterfactual intervention programs}. In Section \ref{sec:semQA}, we discuss the use of domain knowledge and query answering.  We end in Section \ref{sec:last} with some final conclusions. In Appendix \ref{sec:asp} we provide the basics of answer-set programming. In Appendix \ref{sec:example}, we present the complete program for the example, in {\em DLV} code, and its output.

\vspace{-1mm}
\section{Counterfactual Interventions and Explanation Scores}\label{sec:cla}

\vspace{-1mm}
We consider a finite set of features, $\mc{F}$, with each feature $F \in \mc{F}$ having a finite domain, $\nit{Dom}(F)$, where $F$, as a function, takes its values. The features are applied to entities $\e$ that belong to a  population $\mc{E}$. Actually, we identify the entity $\e$ with the record (or tuple) formed by the values the features take on it: \ $\e = \langle F_1(\e), \ldots, F_n(\e)\rangle$.
\ Now, entities in $\mc{E}$ go through a {\em classifier}, $C$, that returns {\em labels} for them. We will assume the classifier is binary, e.g. the  labels could be \ $1$ or $0$.

\begin{figure}[h]
\centerline{\includegraphics[width=4cm]{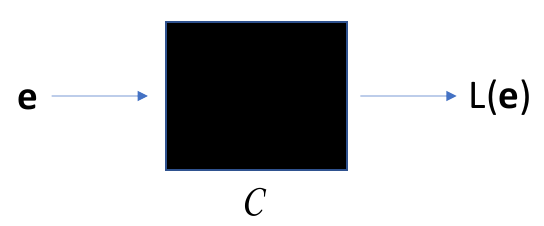}}
\vspace{-3mm}\caption{A black-box classifier} \label{fig:bb}
\end{figure}

In Figure \ref{fig:bb}, we have a classifier receiving as input an entity, $\e$. It returns as an output a label, $L(\e)$, corresponding to the classification of input $\e$.
In principle, we could see $\mc{C}$ as a black-box, in the sense that only by direct interaction with it,  we have access to its input/output relation. That is, we may have no access to the mathematical classification model inside $\mc{C}$.

\vspace{-4mm}

\vspace{-6mm}  The entity $\e$ could represent a client requesting a loan from a financial institution. The classifier of the latter, on the basis of $\e$'s feature values (e.g. for $\mbox{\sf EdLevel}$, $\mbox{\sf Income}$, $\mbox{\sf Age}$, etc.)  assigns the label $1$, for rejection. An explanation may be  requested by the client. Explanations like this could be expected from any kind of classifier. It could be an explicit classification model, e.g. a classification tree or \ a {logistic regression model}. In these cases, we might be in a better position to give an explanation, because we can inspect the internals of the model \cite{rudin}. However, we can find  ourselves in the ``worst-case scenario" in which we do not have access to the internal model. That is, we are confronted to a black-box classifier, and we still have to provide explanations.

An approach to explanations that has become popular, specially in the absence of the model, assigns numerical {\em scores} to the feature values for an entity, trying to answer the question about which   of the feature  values contribute the most to the received label.

\begin{example} \label{ex:NB} We reuse a popular example from \cite{mitchell}.  The set of features is $\mc{F} = \{\mathsf{Outlook}, \msf{Temperature}, \mathsf{Humidity}, \mathsf{Wind}\}$, with $\nit{Dom}(\mathsf{Outlook}) =$ $\{\mathsf{sunny}, \mathsf{overcast},$
$\mathsf{rain}\}$, $\nit{Dom}(\mathsf{Temperature}) = \{\msf{high}, \msf{medium}, \msf{low}\}$, $\nit{Dom}(\mathsf{Humidity}) =$ $\{\mathsf{high}, \mathsf{normal}\}$, $\nit{Dom}(\mathsf{Wind}) =$ $\{\mathsf{strong},$ $ \mathsf{weak}\}$. \ We will always use this order for the features.

Now, assume we have a classifier, $\mc{C}$, that allows us to decide if we play tennis (label $\msf{yes}$) or not (label $\msf{no}$) under a given combination of weather features. A concrete {\em naive-Bayes classifier} will be given in Section  \ref{sec:NB}.
 For example, a particular
 weather entity  has a value for each of the features, e.g. $\e = \nit{ent}(\msf{rain},\msf{high}, \msf{normal}, \msf{weak})$. We want to decide about playing tennis or not under the wether conditions represented by $\e$. \boxtheorem
\end{example}

\ignore{

\begin{center}\begin{tabular*}{7cm}{|c|}\cline{1-1}\\
\phantom{o} \hspace{6cm} \phantom{o}\\
\phantom{o}\\
\phantom{o}\\
\phantom{o}\\
\phantom{o}\\
\phantom{o}\\
\phantom{o}\\
\phantom{o}\\
\phantom{o}\\
\cline{1-1}
\end{tabular*}
\end{center}

\vspace{-5cm}
\begin{figure}[h]
\begin{center}
\includegraphics[width=5.8cm]{decTreeMitchell.pdf}
\caption{A Decision Tree}\label{fig:tree}
\end{center}
\end{figure}
}

Score-based methodologies are sometimes based on {\em counterfactual interventions}: {\em What would happen with the label if we change this particular value, leaving the others fixed?} Or the other way around: {\em What if we leave this value fixed, and change the others?} The resulting labels from these counterfactual interventions can be aggregated in different ways, leading to a score for the feature value under inspection.

Let us illustrate these questions by using the entity $\e$ in the preceding example. If we use the {\em naive-Bayes classifier}  with  entity $\e$, we obtain the label $\msf{yes}$  (c.f. Section \ref{sec:NB}). In order to detect and quantify the relevance (technically, the responsibility) of a feature value in $\e = \nit{ent}(\msf{rain},\msf{high}, \underline{\msf{normal}}, \msf{weak})$, say, of feature $\msf{Humidity}$ (underlined), we {\em hypothetically intervene} its value. In this case, if we change it from $\msf{normal}$ to $\msf{high}$, we obtain a new entity $\e' = \nit{ent}(\msf{rain},\msf{high},\msf{high},\msf{weak})$. If we input this entity $\e'$ into the classifier, we now obtain the label $\msf{no}$. We say that $\e'$ is a {\em counterfactual version} of $\e$.

This change of label is an indication that the original feature value for $\msf{Humidity}$ is indeed relevant for the original classification. Furthermore, the fact that it is good enough to change only this individual value is an indication of its strength. If, to change the label, we also had to change other values together with that for $\msf{Humidity}$, its strength would be lower.
\ In Section \ref{sec:xresp}, we  revisit a particular {\em responsibility score}, $\xresp$, which captures this intuition, and  can be applied with  black-box or open models.

\vspace{-1mm}
\section{A Naive-Bayes Classifier}\label{sec:NB}

\vspace{-2mm}
\begin{multicols}{2}

\begin{tabular}{|c|}\cline{1-1}\\
\phantom{o} \hspace{4.5cm} \phantom{o}\\
\phantom{o}\\
\phantom{o}\\
\phantom{o}\\
\phantom{o}\\
\phantom{o}\\
\phantom{o}\\
\cline{1-1}
\end{tabular}

\vspace{-3.2cm}

\hspace*{2mm}\includegraphics[width=4.8cm]{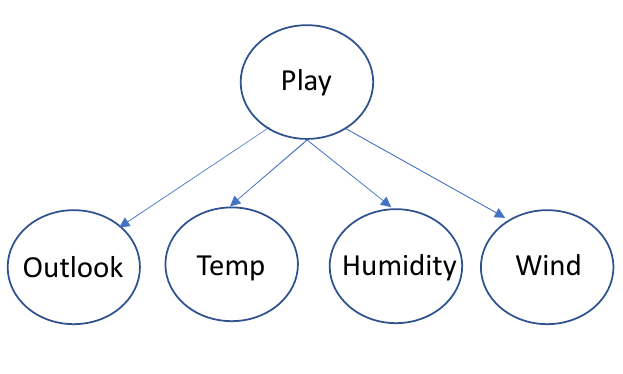}

\phantom{ooo}

\phantom{ooo}

\phantom{ooo}

{\scriptsize
\begin{tabular}{|cccc|c|} \hline
{\sf Outlook} &{\sf Temperature} &{\sf Humidity} & {\sf Wind} & {\sf Play}\\ \hline
{\sf sunny} &{\sf high} &{\sf high} &{\sf weak} &{\sf no}\\
{\sf sunny} &{\sf high} &{\sf high}& {\sf strong}& {\sf no}\\
{\sf overcast}& {\sf high}& {\sf high}& {\sf weak}& {\sf yes}\\
{\sf rain} &{\sf medium} &{\sf high} &{\sf weak}& {\sf yes}\\
{\sf rain} &{\sf low} &{\sf normal} &{\sf weak} &{\sf yes}\\
{\sf rain} &{\sf low}& {\sf normal} &{\sf strong}& {\sf no}\\
{\sf overcast} &{\sf low}& {\sf normal}& {\sf strong} &{\sf yes}\\
{\sf sunny}& {\sf medium} &{\sf high}& {\sf weak}& {\sf no}\\
{\sf sunny}& {\sf low}& {\sf normal}& {\sf weak}& {\sf yes}\\
{\sf rain}& {\sf medium}& {\sf normal}& {\sf weak}& {\sf yes}\\
{\sf sunny}& {\sf medium}& {\sf normal}& {\sf strong}& {\sf yes}\\
{\sf overcast}& {\sf medium}& {\sf high}& {\sf strong}& {\sf yes}\\
{\sf overcast}& {\sf high}& {\sf normal}& {\sf weak}& {\sf yes}\\
{\sf rain}& {\sf medium}& {\sf high}& {\sf strong}& {\sf no}\\ \hline
\end{tabular} }
\end{multicols}

\vspace{-1mm}
\begin{example} \label{ex:NB2} (example \ref{ex:NB} cont. ) We now we build a naive-Bayes classifier for the binary variable $\sf{Play}$, about playing tennis or not. A Bayesian network, that is the basis for this classifier, is shown right here above (left). In addition to the network structure, we have to assign probability distributions to the nodes in it. These distributions are learned from the training data in the table (right).

 In this case, the features stochastically depend on the output variable $\msf{Play}$, and are independent from each other given the output. To fully specify the network, we need the  absolute distribution for the top node; and the conditional distributions for the lower nodes.

These are the distributions inferred from the frequencies in the training data:

\vspace{1mm}
{\footnotesize \begin{center}\begin{tabular}{|l|l|} \hline
$P(\msf{Play} = \msf{yes}) = \frac{9}{14}$  & $P(\msf{Play} = \msf{no}) = \frac{5}{14}$\\ \hline
$P(\msf{Outlook} = \msf{sunny}|\msf{Play} = \msf{yes}) = \frac{2}{9}$ &
$P(\msf{Outlook} = \msf{sunny}|\msf{Play} = \msf{no}) = \frac{3}{5}$\\
$P(\msf{Outlook} = \msf{overcast}|\msf{Play} = \msf{yes}) = \frac{4}{9}$ &
$P(\msf{Outlook} = \msf{overcast}|\msf{Play} = \msf{no})  =0$\\
$P(\msf{Outlook} = \msf{rain}|\msf{Play} = \msf{yes}) = \frac{3}{9}$ &
$P(\msf{Outlook} = \msf{rain}|\msf{Play} = \msf{no}) = \frac{2}{5}$\\
$P(\msf{Temp} = \msf{high}|\msf{Play} = \msf{yes}) = \frac{2}{9}$ &
$P(\msf{Temp} = \msf{high}|\msf{Play} = \msf{no}) = \frac{2}{5}$\\
$P(\msf{Temp} = \msf{medium}|\msf{Play} = \msf{yes}) = \frac{4}{9}$ &
$P(\msf{Temp} = \msf{medium}|\msf{Play} = \msf{no}) = \frac{2}{5}$\\
$P(\msf{Temp} = \msf{low}|\msf{Play} = \msf{yes}) = \frac{3}{9}$&
$P(\msf{Temp} = \msf{low}|\msf{Play} = \msf{no}) = \frac{1}{5}$\\
$P(\msf{Humidity} = \msf{high}|\msf{Play} = \msf{yes}) = \frac{3}{9}$&
$P(\msf{Humidity} = \msf{high}|\msf{Play} = \msf{no}) = \frac{4}{5}$\\
$P(\msf{Humidity} = \msf{normal}|\msf{Play} = \msf{yes}) = \frac{6}{9}$&
$P(\msf{Humidity} = \msf{normal}|\msf{Play} = \msf{no}) = \frac{1}{5}$\\
$P(\msf{Wind} = \msf{strong}|\msf{Play} = \msf{yes}) = \frac{3}{9}$&
$P(\msf{Wind} = \msf{strong}|\msf{Play} = \msf{no}) = \frac{3}{5}$\\
$P(\msf{Wind} = \msf{weak}|\msf{Play} = \msf{yes}) = \frac{6}{9}$&
$P(\msf{Wind} = \msf{weak}|\msf{Play} = \msf{no}) = \frac{2}{5}$\\ \hline
\end{tabular}\end{center} }

\vspace{1mm}

We can use them to decide, for example, about playing or not with the following input data: \
$\msf{Outlook}=\msf{rain}, \msf{Temp}=\msf{high}, \msf{Humidity}=\msf{normal},
\msf{Wind}=\msf{weak}$. \ If we keep this order of the features, we are classifying the weather entity $\e = \langle  \msf{rain}, \msf{high}, \msf{normal},
\msf{weak}\rangle$. \
This is done by determining the maximum probability between the two probabilities:
\begin{eqnarray}
P(\msfs{Play} &=& \msfs{yes}|\msfs{Outlook}=\msfs{rain}, \msfs{Temp}=\msfs{high}, \msfs{Humidity}=\msfs{normal},
\msfs{Wind}=\msfs{weak} ), \label{eq:num1}\\
P(\msfs{Play} &=& \msfs{no}| \msfs{Outlook}=\msfs{rain}, \msfs{Temp}=\msfs{high}, \msfs{Humidity}=\msfs{normal},
\msfs{Wind}=\msfs{weak}).\label{eq:num2}
\end{eqnarray}
Now, for each of the probabilities of the form $P(\msf{P}|\msf{O},\msf{T},\msf{H},\msf{W})$ it holds:
\begin{equation}P(\msfs{P}|\msfs{O},\msfs{T},\msfs{H},\msfs{W}) = \frac{P(\msfs{P},\msfs{O},\msfs{T},\msfs{H},\msfs{W})}{P(\msfs{O},\msfs{T},\msfs{H},\msfs{W})}
 = \frac{P(\msfs{O}|\msfs{P}) P(\msfs{T}|\msfs{P})P(\msfs{H}|\msfs{P})P(\msfs{W}|\msfs{P}) P(\msfs{P})}{\sum_{\msfs{P}} P(\msfs{O}|\msfs{P}) P(\msfs{T}|\msfs{P})P(\msfs{H}|\msfs{P})P(\msfs{W}|\msfs{P}) P(\msfs{P})}. \label{eq:fla}
 \end{equation}
In particular, the numerators for (\ref{eq:num1}) and (\ref{eq:num2}) become, resp.:
{\small
\begin{eqnarray}
&&P(\msfs{Outlook}=\msfs{rain}|\msfs{Play} = \msfs{yes}) P( \msfs{Temp}=\msfs{high}|\msfs{Play} = \msfs{yes}) P(\msfs{Humidity}=\msfs{normal}|\msfs{Play} = \msfs{yes}) \times \nonumber \\
&&\times P(\msfs{Wind}=\msfs{false}|\msfs{Play} = \msfs{yes}) P(\msfs{Play} = \msf{yes}) = \frac{3}{9} \frac{2}{9} \frac{6}{9}\frac{6}{9}\frac{9}{14} = \frac{4}{189}, \label{eq:1}\\
&&P(\msfs{Outlook}=\msfs{rain}|\msfs{Play} = \msfs{no}) P( \msfs{Temp}=\msfs{high}|\msfs{Play} = \msfs{no}) P(\msfs{Humidity}=\msfs{normal}|\msfs{Play} = \msfs{no}) \times \nonumber\\
&&\times P(\msfs{Wind}=\msfs{false}|\msfs{Play} = \msfs{no}) P(\msfs{Play} = \msfs{no}) = \frac{2}{5} \frac{2}{5}\frac{1}{5}\frac{2}{5}\frac{5}{14} = \frac{4}{875}. \label{eq:2}
\end{eqnarray} }
The denominator for both cases is the marginal probability, i.e. \ $\frac{ 4}{189} + \frac{4}{875}$. Then, it is good enough to compare (\ref{eq:1}) and (\ref{eq:2}). Since the
former is larger, the decision (or classification) becomes:  \ $\msf{Play} = \msf{yes}$. \boxtheorem
\end{example}

\vspace*{-3mm}
\section{The $\mbox{\sf x-Resp}$ Score}\label{sec:xresp}

\vspace{-2mm}
Assume that an entity $\e$ has received the label $1$ by the classifier $\mc{C}$, and we want to explain this outcome by assigning
numerical scores to $\e$'s feature values, in such a way, that a higher score for a feature value reflects that it has been important for the outcome. We do this now using the  $\mbox{\sf x-Resp}$ score, whose definition we motivate below by means of an example.  The $\xresp$ score as defined below is not restricted to- but  more suitable for  binary features, i.e. that take the values $\mbox{\sf true}$ or    $\mbox{\sf false}$ (or $1$ and $0$, resp.). The generalization in \cite{tplp} is more appropriate for multi-valued features. C.f. Section \ref{sec:last} for a discussion, and \cite{rr20,tplp} for more details.

 \begin{example} In Figure \ref{fig:cl}, the black box is classifier $\mc{C}$. An entity $\e$ has gone through it obtaining label $1$, shown in the first row in the figure. We want to assign a score to the feature value $\mathbf{x}$ for a feature $F \in \mc{F}$. \ We proceed, counterfactually, changing the value   $\mathbf{x}$ into $\mathbf{x}'$, obtaining a counterfactual version $\e_1$ of $\e$. We classify $\e_1$, and we still get the outcome $1$ (second row). In between, we may counterfactually change other feature values, $\mathbf{y}, \mathbf{z}$ in $\e$, into $\mathbf{y}', \mathbf{z}'$, but keeping $\mathbf{x}$, obtaining entity $\e_2$, and the outcome does not change (third row). However, if we change in $\e_2$, $\mathbf{x}$ into $\mathbf{x}'$, the outcome does change (fourth row).

\vspace{-8mm}
\begin{figure}[h]
\begin{center}
\includegraphics[width=8.5cm]{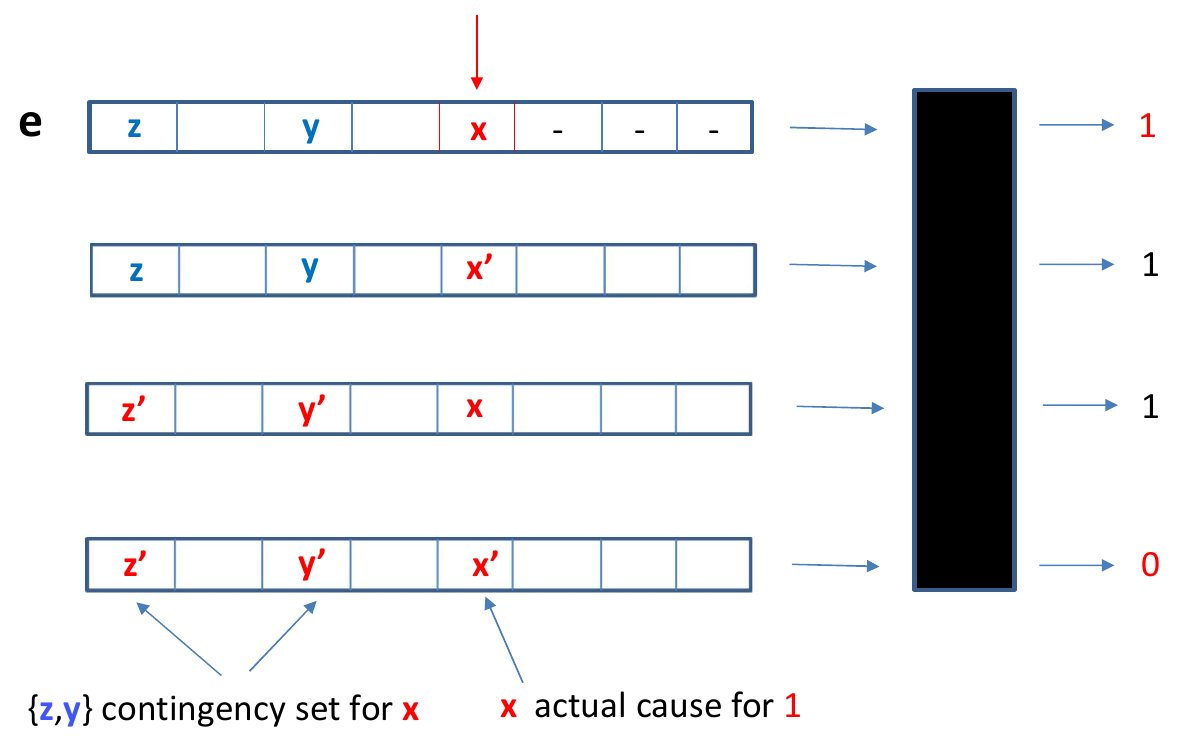}

\vspace{-4mm}\caption{Classified entity and its counterfactual versions}\label{fig:cl}
\end{center}
\end{figure}

\vspace{-8mm}
This shows that the value $\mathbf{x}$ is relevant for the original output, but, for this outcome,  it needs company, say of the feature values $\mathbf{y}, \mathbf{z}$ in $\e$. According to {\em actual causality}, we can say that the feature value $\mathbf{x}$ in $\e$ is an {\em actual cause} for the classification, that needs a {\em contingency set} formed by the values $\mathbf{y}, \mathbf{z}$ in $\e$. In this case, the contingency set has size $2$. If we found a contingency set for $\mathbf{x}$ of size $1$ in $\e$, we would consider $\mathbf{x}$ even more relevant for the output.
\boxtheorem
 \end{example}

On this basis, we can define \cite{rr20,tplp}, for a feature value $\mathbf{x}$ in $\e$: \ (a) ${\mathbf{x}}$ \ is a {\em counterfactual explanation} for \ ${L(\mathbf{e}) =1}$ \ if \ ${L(\mathbf{e}\frac{{\mathbf{x}}}{\mathbf{x}^\prime}) = 0}$, \ for some ${\mathbf{x}^\prime \in \nit{Dom}(F)}$ (the domain of feature $F$). (Here we use the common notation $\mathbf{e}\frac{{\mathbf{x}}}{\mathbf{x}^\prime}$ for the entity obtained by replacing $\mathbf{x}$ by $\mathbf{x}'$ in $\mathbf{e}$.) \ (b)
 ${\mathbf{x}}$ \ is an {\em actual explanation} for \ ${L(\mathbf{e}) =1}$ \ if there is a contingency set  of values ${\mathbf{Y}}$ in ${\mathbf{e}}$, with ${{\mathbf{x}} \notin \mathbf{Y}}$, and new values ${\mathbf{Y}^\prime \cup \{\mathbf{x}^\prime\}}$, such that \ ${L(\mathbf{e}\frac{\mathbf{Y}}{\mathbf{Y}^\prime}) = 1}$  and \ ${L(\mathbf{e}\frac{{\mathbf{x}}\mathbf{Y}}{\ \mathbf{x}^\prime\mathbf{Y}^\prime}) = 0}$.
\  We say that $\mathbf{Y}$ is {\em minimal} if there is no $\mathbf{Y}'$ with $\mathbf{Y}' \subsetneqq  \mathbf{Y}$, that is also a contingency set for $\mathbf{x}$ in $\e$. Similarly, $\mathbf{Y}$ is {\em miminum} contingency set if it is a a minimum-size contingency set for $\mathbf{x}$ in $\e$.

Contingency sets may come in sizes from $0$ to $n-1$ for feature values in records of length $n$. Accordingly, we can define for the actual cause $\mbf{x}$ in $\e$: If ${\mbf{Y}}$ is a minimum contingency set for $\mbf{x}$, $\mbox{\sf x-Resp}(\e,\mbf{x}) := \frac{1}{1 + |\mbf{Y}|}$; \ and as $0$ when $\mbf{x}$ is not an actual cause. \ \ (C.f. Section \ref{sec:last} for the $\resp$ score that generalizes $\xresp$.)

{\em We will reserve the notion of {\em counterfactual explanation} for (or {\em counterfactual version} of) an input entity $\e$ for any entity $\e'$ obtained from $\e$ by modifying feature values in $\e$ and that leads to a different label, i.e. $L(\e) \neq L(\e')$. Notice that from such an $\e'$ we can read off actual causes for $L(\e)$ as feature values, and contingency sets for those actual causes. It suffices to compare $\e$ with $\e'$.}
\ignore{
\red{\comgr{El parrafo de arriba es completo en letra cursiva, o sólo {\em counterfactual explanation} y {\em counterfactual version} ??}}
}

In Section \ref{sec:cips} we give a detailed example that illustrates these notions, and also shows the use of ASPs for the specification and computation of counterfactual versions of a given entity, and the latter's $\mbox{\sf x-Resp}$ score.

\vspace{-2mm}
\section{Counterfactual-Intervention Programs}\label{sec:cips}

\vspace{-1mm}

Together with illustrating the notions introduced in Section \ref{sec:xresp}, we will introduce, by means of an example,  {\em Counterfactual Intervention Programs} (CIPs). The program corresponds to the naive-Bayes classifier presented in Section \ref{sec:NB}.

CIPs are {\em answer-set programs} (ASPs) that specify the counterfactual versions of a given entity, and also, if so desired, only the {\em maximum-responsibility} counterfactual explanations, i.e. counterfactual versions that lead to a maximum $\mbox{\sf x-Resp}$ score. \ (C.f. \cite{tplp} for many more details and examples with {\em decision trees} as classifiers.).

\vspace{-2mm}
\begin{example} \label{ex:nb2}  \ (examples \ref{ex:NB} and \ref{ex:NB2} cont.) \ We present now the CIP in {\em DLV-Complex} notation.  \ Since the program specifies and applies counterfactual  changes of attribute values, we have to indicate when an intervention is applied, when an entity may still be subject to additional interventions, and when a final version of an entity has been reached, i.e. the label has been changed. To achieve this, the program uses annotation constants \verb+o+, for ``original entity", \verb+do+, for ``do a counterfactual intervention" (a single change of feature value), \verb+tr+, for ``entity in transition", and \verb+s+, for ``stop, the label has changed".

We will explain the program along the way, as we present it, and with additional explanations as comments written directly in the {\em DLV} code. We will keep the most relevant parts of the program. The complete program can be found in Appendix \ref{sec:example}.

The absolute and conditional probabilities will be given as facts of the {\em DLV} program. They  are represented as percentages, because {\em DLV} handles operations with integer numbers. The conditional probabilities are atoms of the form \verb+p_f_c(feature value, play outcome, prob\%)+, with ``\verb+f+" suggesting the feature name, and ``\verb+c+", that it is a conditional probability. For example, \linebreak \verb+p_h_c(normal, yes, 67)+ is the conditional probability (of $67\%$) of $\msf{Humidity}$ being $\msf{normal}$  given that $\msf{Play}$  takes value $\msf{yes}$. Similarly, this is an absolute probability for $\msf{Play}$: \ \verb+p(yes, 64)+.

The program has as facts also the contents of the  domains. They are of the form \ \verb+dom_f(feature value)+, with ``\verb+f+" suggesting the feature name again, e.g. \verb+dom_h(high)+, for $\msf{Humidity}$. Finally, among the facts we find the original entity  that will be intervened by means of the CIP. \ In this case, as in Example \ref{ex:NB}, \verb+ent(e,rain,high,normal,weak,o)+, where constant \verb+e+ is an entity identifier (eid), and \verb+o+ is the annotation constant. This entity gets label $\msf{yes}$, i.e.  $\msf{Play} ~=~ \msf{yes}$. Through interventions, we expect the label to become $\msf{no}$, i.e. $\msf{Play} = \msf{no}$.

 Aggregation functions over sets will be needed later in the program, to build {\em contingency sets} (c.f. Section \ref{sec:xresp}). So, we use {\em DLV-Complex} that supports this functionality. ``List and Sets" has to be specified at the beginning of the program, together with the maximum integer value. This is the first part of the CIP, showing the facts: \ (as usual, words starting with lower case are constants; whereas with upper case, variables)

\vspace{-1mm}
{\footnotesize
\begin{verbatim}
 % DLV-COMPLEX    #include<ListAndSet>   #maxint = 100000000.
 % domains:
    dom_o(sunny). dom_o(overcast). dom_o(rain). dom_t(high). dom_t(medium).
    dom_t(low). dom_h(high). dom_h(normal). dom_w(strong). dom_w(weak).
 % original entity that gets label 1:
    ent(e,rain,high,normal,weak,o).
 % absolute probabilities for Play (as percentage)
    p(yes, 64). p(no, 36).
 % Outlook conditional probabilities (as percentage)
    p_o_c(sunny, yes, 22). p_o_c(overcast, yes, 45). p_o_c(rain, yes, 33).
    p_o_c(sunny, no, 60). p_o_c(overcast, no, 0). p_o_c(rain, no, 40).
 % Temperature conditional probabilities (as percentage)
    p_t_c(high, yes, 22). p_t_c(medium, yes, 45). p_t_c(low, yes, 33).
    p_t_c(high, no, 40). p_t_c(medium, no, 40). p_t_c(low, no, 20).
 % Humidity conditional probabilities (as percentage)
    p_h_c(normal, yes, 67). p_h_c(high, yes, 33).
    p_h_c(normal, no, 20). p_h_c(high, no, 80).
 % Wind conditional probabilities (as percentage)
    p_w_c(strong, yes, 33). p_w_c(weak, yes, 67).
    p_w_c(strong, no, 60). p_w_c(weak, no, 40).
\end{verbatim}}

\vspace{-1mm} The classifier will compute posterior probabilities for $\msf{Play}$ according to equations (\ref{eq:num1}) and (\ref{eq:num2}) in Section \ref{sec:NB}. Next, they
are compared, and the largest determines the label. As we can see from  equation (\ref{eq:fla}), the denominator is irrelevant for this comparison. So, we  need only the numerators. They are specified by means of a predicate of the form \verb+pb_num(E,O,T,H,W,V,Fp)+, where the arguments stand for: eid, (values for) $\msf{Outlook}$,  $\msf{Temp}$, $\msf{Humidity}$, $\msf{Wind}$ and $\msf{Play}$, resp.; and the probability as a percentage.
\ The CIP has to specify predicate \verb+pb_num(E,O,T,H,W,V,Fp)+. That part of the program is not particularly interesting, and looks somewhat  cumbersome due to the combination of simple arithmetical operations with probabilities. \ignore{The reason is that, in order to reach the value for its last argument, \verb+Fp+, one has to perform four steps of arithmetical operations with probabilities (as in the numerator of (\ref{eq:fla})) given as percentages. } C.f. the program  in Appendix \ref{sec:example}.

Next, we have to specify the transition annotation constant \verb+tr+, that is used  in rule bodies below. It indicates that we are using  an entity that is in transition. This annotation is specified as follows:

{\footnotesize
\begin{verbatim}
 % transition rules: the initial entity or one affected by an intervention
   ent(E,O,T,H,W,tr) :- ent(E,O,T,H,W,o).
   ent(E,O,T,H,W,tr) :- ent(E,O,T,H,W,do).
   \end{verbatim}}
\vspace{-4mm}   Now we have to specify the classifier, or better, the classification criteria, appealing to predicate \verb+pb_num(E, O, H, W, V, Fp)+. More precisely, we have to compare \verb+Fp+ for Play value \verb+yes+, denoted \verb+Fyes+, with \verb+Fp+ for Play value \verb+no+, denoted \verb+Fno+. If the former is larger, we obtain label $\msf{yes}$; otherwise label $\msf{no}$:

 {\footnotesize
\begin{verbatim}
 % spec of the classifier
   cls(E,O,T,H,W,yes) :- ent(E,O,T,H,W,tr),  pb_num(E,O,T,H,W,yes,Fyes),
                         pb_num(E,O,T,H,W,no,Fno), Fyes >= Fno.
   cls(E,O,T,H,W,no)  :- ent(E,O,T,H,W,tr),  pb_num(E,O,T,H,W,yes,Fyes),
                         pb_num(E,O,T,H,W,no,Fno), Fyes < Fno.
\end{verbatim}}
Notice the use of annotation constant \verb+tr+ in the body, because we will be classifying entities that are in transition.
\ Next, the CIP specifies all the one-step admissible counterfactual interventions on entities with label $\msf{yes}$, which produces entities in transition. This disjunctive rule is the main rule.
{\footnotesize
\begin{verbatim}
 % counterfactual rule: alternative single-value changes
   ent(E,Op,T,H,W,do) v ent(E,O,Tp,H,W,do) v
   ent(E,O,T,Hp,W,do) v ent(E,O,T,H,Wp,do) :- ent(E,O,T,H,W,tr),
        cls(E,O,T,H,W,yes), O != Op, T != Tp, H!= Hp, W!= Wp,
        chosen_o(O,T,H,W,Op), chosen_t(O,T,H,W,Tp), chosen_h(O,T,H,W,Hp),
        chosen_w(O,T,H,W,Wp), dom_o(Op), dom_t(Tp), dom_h(Hp), dom_w(Wp).
\end{verbatim}}

\ignore{
   chosen_o(O,T,H,W,U) :- ent(E,O,T,H,W,tr), cls(O,T,H,W,1), dom_o(U),
                         U != O, not diffchoice_o(O,T,H,W,U).
   diffchoice_o(O,T,H,W,U) :- chosen_o(O,T,H,W, Up), U != Up, dom_o(U).
   chosen_t(O,T,H,W,U) :- ent(E,O,T,H,W,tr), cls(O,T,H,W,1), dom_t(U),
                         U != T, not diffchoice_t(O,T,H,W,U).
   diffchoice_t(O,T,H,W,U) :- chosen_t(O,T,H,W, Up), U != Up, dom_t(U).
   chosen_h(O,T,H,W,U) :- ent(E,O,T,H,W,tr), cls(O,T,H,W,1), dom_h(U),
                         U != H, not diffchoice_h(O,T,H,W,U).
   diffchoice_h(O,T,H,W,U) :- chosen_h(O,T,H,W, Up), U != Up, dom_h(U).
   chosen_w(O,T,H,W,U) :- ent(E,O,T,H,W,tr), cls(O,T,H,W,1), dom_w(U),
                         U != W,  not diffchoice_h(O,T,H,W,U).
   diffchoice_w(O,T,H,W,U) :- chosen_h(O,T,H,W, Up), U != Up, dom_w(U).
   }
Here we are using predicates \verb+chosen+, one for each of the four features. For example, \verb+chosen_h(O,T,H,W,Hp)+ ``chooses" for each combination of values, \verb+O,T,H,W+ for $\msf{Outlook}$,
   $\msf{Temp}$, $\msf{Humidity}$, and $\msf{Wind}$, a unique (and new) value \verb+Hp+ for feature $\msf{Humidity}$, and that value is taken from its domain \verb+dom_h+. Through an intervention, that value \verb+Hp+ replaces the original value \verb+H+, as one of the four possible value changes that are indicated in the rule head.

The semantics of ASPs makes only one of the possible disjuncts in the head true (unless forced otherwise by other rules in the program, which does not happen with CIPs). The \verb+chosen+ predicates can be specified in a generic manner \cite{zaniolo}. Here, we skip their specification, but they can be found in Appendix \ref{sec:example}.

In order to avoid going back to the original entity through counterfactual interventions, we may impose a {\em hard program constraint} \cite{leone}. These constraints are rules with empty head, which capture a negation. They have the effect of discarding the models where the body becomes true. In this case:
      {\footnotesize
\begin{verbatim}
 % not going back to initial entity
       :- ent(E,O,T,H,W,do), ent(E,O,T,H,W,o).
\end{verbatim} }
Next, we stop performing interventions when we switch the label to \verb+no+, which introduces the annotation \verb+s+:
     {\footnotesize
\begin{verbatim}
 % stop when label has been changed:
   ent(E,O,T,H,W,s) :- ent(E,O,T,H,W,do), cls(E,O,T,H,W,no).
   \end{verbatim} }
Finally, we introduce an extra program constraint, to avoid computing models where the original entity never changes label. Those models will not contain the original eid with annotation \verb+s+:
  {\footnotesize
\begin{verbatim}
   % extra constraint avoiding models where label does not change
       :- ent(E,O,T,H,W,o), not entAux(E).
   % auxiliary predicate to avoid unsafe negation right above
     entAux(E) :- ent(E,O,T,H,W,s).
\end{verbatim}}
The rest of the program uses counterfactual interventions to collect individual changes (next rules), sets of them, cardinalities of those sets, etc.
  {\footnotesize
\begin{verbatim}
 % collecting changed values for each feature:
   expl(E,outlook,O)  :- ent(E,O,T,H,W,o), ent(E,Op,Tp,Hp,Wp,s), O != Op.
   expl(E,temp,T)     :- ent(E,O,T,H,W,o), ent(E,Op,Tp,Hp,Wp,s), T != Tp.
   expl(E,humidity,H) :- ent(E,O,T,H,W,o), ent(E,Op,Tp,Hp,Wp,s), H != Hp.
   expl(E,wind,W)     :- ent(E,O,T,H,W,o), ent(E,Op,Tp,Hp,Wp,s), W != Wp.
\end{verbatim}}
With them, we will obtain, for example, the atom \verb+expl(e,humidity,normal)+ in some of the models of the program, because there is a counterfactual entity that changes normal humidity into high (c.f. Section \ref{sec:cla}). The atom indicates that original value \verb+normal+ for \verb+humidity+ is part of an explanation for entity \verb+e+.
\ignore{
  {\footnotesize
\begin{verbatim}
 % collecting changed values for each feature:
   expl(E,Ou,O) :- entSchema(Ou,Te,Hu,Wi), ent(E,O,T,H,W,o),
                   ent(E,Op,Tp,Hp,Wp,s), O != Op.
   expl(E,Te,T) :- entSchema(Ou,Te,Hu,Wi), ent(E,O,T,H,W,o),
                   ent(E,Op,Tp,Hp,Wp,s), T != Tp.
   expl(E,Hu,H) :- entSchema(Ou,Te,Hu,Wi), ent(E,O,T,H,W,o),
                   ent(E,Op,Tp,Hp,Wp,s), H != Hp.
   expl(E,Wi,W) :- entSchema(Ou,Te,Hu,Wi), ent(E,O,T,H,W,o),
                   ent(E,Op,Tp,Hp,Wp,s), W != Wp.
\end{verbatim}}}
Contingency sets for a feature value are obtained  with the rules below. Since we keep everywhere the eid, it is good enough to collect the names of the features whose values are changed. For this we use predicate
\verb+cont(E,U,S)+. Here, \verb+U+ is a feature (with changed value), \verb+S+ is the {\em set of all} \ feature names whose values are changed together with that for \verb+U+. These sets are build using the built-in set  functions of {\em DLV-Complex}. Similarly with the built-in set membership check.

\vspace{-1mm}
{\footnotesize
\begin{verbatim}
 % building  contingency sets
   cause(E,U)       :- expl(E,U,X).
   cauCont(E,U,I)   :- expl(E,U,X), expl(E,I,Z), U != I.
   preCont(E,U,{I}) :- cauCont(E,U,I).
   preCont(E,U,#union(Co,{I})) :- cauCont(E,U,I), preCont(E,U,Co),
                                  not #member(I,Co).
   cont(E,U,Co)     :- preCont(E,U,Co), not HoleIn(E,U,Co).
   HoleIn(E,U,Co)   :- preCont(E,U,Co), cauCont(E,U,I), not #member(I,Co).
   tmpCont(E,U)     :- cont(E,U,Co), not #card(Co,0).
   cont(E,U,{})     :- cause(E,U), not tmpCont(E,U).
   \end{verbatim}}

\vspace{-5mm}   The construction is such that one keeps adding contingency features, using pre-contingency sets,  until there is nothing else to add. In this way the contingency sets contain all the features that have to be changed with the one at hand \verb+U+. \ For example,  in one of the models we will find the atom \verb+cont(e,humidity,{})+, meaning that a change of the humidity value alone, i.e. with empty contingency set, suffices to switch the label. \ Each counterfactual version of  entity $\e$ will be represented by a model of the program. Due to model minimality, the associated set of changes of feature values that accompany a counterfactual change of feature value, say $\mbf{x}$ in $\e$, will correspond to a {\em minimal}, but not necessarily minimum, contingency set $\mbf{Y}$ for $\mbf{x}$ in $\e$ (c.f. Section \ref{sec:xresp}).

   The generation of contingency sets is now useful for the computation of the inverse of the $\mbox{\sf x-Resp}$ score. For this we can use the built-in set-cardinality operation \verb+#card(S,M)+ of {\em DLV-Complex}. Here, \verb+M+ is the cardinality of  set \verb+S+. The score will be the result of adding $1$ to the cardinality \verb+M+ of a contingency set \verb+S+:

{\footnotesize
\begin{verbatim}
 % computing the inverse of x-Resp
   invResp(E,U,R) :- cont(E,U,S), #card(S,M), R = M+1, #int(R).
\end{verbatim}}
For each counterfactual version of $\e$, as represented by a model of the program, we will obtain a {\em local} $\xresp$ score. So, a particular feature value, \verb+U+, may have several local $\xresp$ scores in different models of the program.
For example, in the model corresponding to the change of humidity (and nothing else) we will get the atom \verb+invResp(e,humidity,1)+. Finally, full explanations will be of the form \verb+fullExpl(E,U,R,S)+, where \verb+U+ is a feature name, \verb+R+ is its inverse $\xresp$ score, and \verb+S+ is its contingency set (of feature names).
    {\footnotesize
\begin{verbatim}
 % full explanations:
   fullExpl(E,U,R,S) :-  expl(E,U,X), cont(E,U,S), invResp(E,U,R).
\end{verbatim}}
Following with our ongoing example, we will get in one model the atom \verb+fullExpl(e,humidity,1,{})+. \ Additional information, such as the new feature values that lead to the change of label can be read-off from the associated model (examples follow). The original feature values can be recovered via the eid \verb+e+ from the original entity.

If we run the program starting with the original entity, we obtain ten different counterfactual versions of $\e$. They are represented by the ten essentially different stable models of the program, and can be read-off from the atoms with the  annotation \verb+s+, namely:  \ (with value changes underlined)
{\small \begin{enumerate}
\item \verb+ent(e,rain,high,+\ul{high}\verb+,weak,s)+
\item \verb+ent(e,rain,high,+\ul{high}\verb+,+\ul{strong}\verb+,s)+,\ \verb+ent(e,+\ul{sunny}\verb+,high,normal,+\ul{strong}\verb+,s)+,  \newline \verb+ent(e,+\ul{sunny}\verb+,high,+\ul{high}\verb+,weak,s)+
\item \verb+ent(e,rain,+\ul{medium}\verb+,+\ul{high}\verb+,+\ul{strong}\verb+,s)+, \ \verb+ent(e,rain,+\ul{low}\verb+,+\ul{high}\verb+,+\ul{strong}\verb+,s)+,  \newline \verb+ent(e,+\ul{sunny}\verb+,+\ul{low}\verb+,+\ul{high}\verb+,weak,s)+, \verb+ent(e,+\ul{sunny}\verb+,+\ul{medium}\verb+,+\ul{high}\verb+,weak,s)+;
     \item \verb+ent(e,+\ul{sunny}\verb+,+\ul{medium}\verb+,+\ul{high}\verb+,+\ul{strong}\verb+,s)+, \verb+ent(e,+\ul{sunny}\verb+,+\ul{low}\verb+,+\ul{high}\verb+,+\ul{strong}\verb+,s)+.
\end{enumerate}}

\vspace{-1mm}Below we show only three of the obtained models (the others are found in Appendix \ref{sec:example}). In the models we show only the  most relevant atoms, omitting initial facts, intermediate probabilities,  and chosen-related atoms:
{\footnotesize
\begin{verbatim}
 M1 {ent(e,rain,high,normal,weak,o), ent(e,rain,high,normal,weak,tr),
     cls(e,rain,high,normal,weak,yes), ent(e,rain,high,high,weak,do),
     ent(e,rain,high,high,weak,tr), cls(e,rain,high,high,weak,no),
     ent(e,rain,high,high,weak,s), expl(e,humidity,normal),
     cont(e,humidity,{}),invResp(e,humidity,1),fullExpl(e,humidity,1,{})}
 M2 {ent(e,rain,high,normal,weak,o), ent(e,rain,high,high,strong,tr),
     cls(e,rain,high,high,strong,no), ent(e,rain,high,high,strong,s),
     invResp(e,humidity,2), fullExpl(e,humidity,2,{wind}),
     invResp(e,wind,2), fullExpl(e,wind,2,{humidity})}
 M3 {ent(e,rain,high,normal,weak,o), ent(e,sunny,high,normal,strong,tr),
     cls(e,sunny,high,normal,strong,no),ent(e,sunny,high,normal,strong,s),
     invResp(e,outlook,2), fullExpl(e,outlook,2,{wind}), ...}
\end{verbatim}}

\vspace{-1mm}The first model corresponds to our running example. The second model shows that the same change of the previous model accompanied by a change for $\msf{Wind}$ also leads to a change of label. We might prefer the first model. We will take care of this next. The third model shows a different combination of changes: for $\msf{Outlook}$ accompanied by  $\msf{Wind}$. In this model, the original $\msf{Outlook}$ value has $\frac{1}{2}$ as $\xresp$ score.

If we are interested only in those counterfactual entities that are obtained through a minimum number of changes, and then leading to maximum responsibility scores, we can impose
{\em weak program constraints} on the program \cite{leone}. In contrast to hard constraints, as used above, they can be violated by a model of the program. However, only those models where the number of violations is a minimum are kept. In our case, the number of value differences between the original and final entity is minimized:
{\footnotesize
\begin{verbatim}
 % weak constraints to minimize number of changes
   :~ ent(E,O,T,H,W,o), ent(E,Op,Tp,Hp,Wp,s), O != Op.
   :~ ent(E,O,T,H,W,o), ent(E,Op,Tp,Hp,Wp,s), T != Tp.
   :~ ent(E,O,T,H,W,o), ent(E,Op,Tp,Hp,Wp,s), H != Hp.
   :~ ent(E,O,T,H,W,o), ent(E,Op,Tp,Hp,Wp,s), W != Wp.
\end{verbatim}}
Running the program with them, leaves only model \verb+M1+ above, corresponding to the counterfactual entity $\e' = \nit{ent}(\mathsf{rain},\mathsf{high},\mathsf{high},\mathsf{weak})$. This is a maximum-responsibility counterfactual explanation. \boxtheorem
\end{example}

\vspace{-2mm}
\section{Exploiting Domain Knowledge and Query Answering}\label{sec:semQA}

\vspace{-1mm}
CIPs allows for the inclusion of domain knowledge. In our example, describing a particular geographic region, it might be the case that there is never high temperature with a strong wind. Such a combination might not be allowed in counterfactuals, which could be done by imposing the program constraint:
{\footnotesize
\begin{verbatim}
 :- ent(E,_,high,_,strong,tr).
\end{verbatim}}
If we run the program with this constraint, models \verb+M2+ and \verb+M3+ above would be discarded, so as any other where the inadmissible combination appears \cite{blgf}.

In another geographic region, it could be the case that there is a functional relationship between features, for example, between $\msf{Temperature}$ and $\msf{Humidity}\!: \ \msf{high} \mapsto  \msf{normal}, \  \{\msf{medium},\msf{low}\} \mapsto \msf{high}$. \ In this case, from the head of the counterfactual rule, the disjunct \verb+ent(E,O,T,Hp,W,do)+ could be dropped for not representing an admissible counterfactual. Instead, we could add the extra rules:
{\footnotesize
\begin{verbatim}
  ent(E,O,T,normal,W,tr) :- ent(E,O,high,H,W,tr).
  ent(E,O,T,high,W,tr)   :- ent(E,O,medium,H,W,tr).
  ent(E,O,T,high,W,tr)   :- ent(E,O,low,H,W,tr).
\end{verbatim} }

\vspace{-1mm}We can also exploit reasoning, which is enabled by query answering. Actually, the models of the program are implicitly queried, as databases (the models do not have to be returned, only the answers). Under the {\em cautious semantics} we obtain the answers that are true in {\em all} models, whereas under  the {\em brave semantics}, the answers that are true in {\em some} model \cite{leone}. They can be used for different kinds of queries.
\ The query semantics is specified  when calling the program (\verb+naiveBayes.txt+), so as the file containing the query (\verb+queries.txt+):

{\footnotesize
\begin{verbatim}
 \DLV>dlcomplex.exe -nofacts -nofdcheck -brave naiveBayes.txt queries.txt
\end{verbatim} }

If we do not use the weak constraints that minimize the responsibility, and we want the responsibility of feature $\msf{Outlook}$, we can pose the query \verb+Q1+ below under the brave semantics. The same to know if there is an explanation with less than 3 changes (\verb+Q2+):\vspace{-3mm}
{\footnotesize
\begin{verbatim}
                            invResp(e,outlook,R)?                     %Q1
                            fullExpl(E,U,R,S), R<3?                   %Q2
\end{verbatim} } \vspace{-2mm}
\verb+Q1+ returns \verb+2+, \verb+3+, and \verb+4+, then the responsibility for $\msf{Outlook}$ is $\frac{1}{2}$. \verb+Q2+ returns all the full explanations with inverse score \verb+1+ or \verb+2+, e.g. \verb+e,outlook,2,{humidity}+. We can also ask, under the brave semantics, if there is an intervened entity exhibiting the combination of sunny outlook with strong wind, and its label (\verb+Q3+). Or perhaps, all the intervened entities that obtained label \verb+no+ (\verb+Q4+): \vspace{-2mm}
{\footnotesize
\begin{verbatim}
                        cls(E,O,T,H,W,_), O = sunny, W = strong?      %Q3
                        cls(E,O,T,H,W,no)?                            %Q4
\end{verbatim} } \vspace{-2mm}
For \verb+Q3+ we obtain, for example, \ \verb+e,sunny,low,normal,strong,yes+; and for \verb+Q4+, for example  \verb+e,sunny,low,high,strong+. We can ask, under the {\em cautions semantics}, whether the wind does not change under every counterfactual version:
{\footnotesize
\begin{verbatim}
                    ent(e,_,_,_,Wp,s), ent(e,_,_,_,W,o), W = Wp?      %Q5
\end{verbatim} }

\vspace{-2mm} We obtain the empty output, meaning $\msf{Wind}$ is indeed changed in at least one counterfactual version (i.e. stable model). In fact, the same query under the {\em brave semantics} returns the records where $\msf{Wind}$ remained unchanged, e.g. \verb+rain,high,high,weak+, along with the original entity \verb+rain,high,normal,weak+. \ignore{Outputs are built by filling the variables that are not strictly specified in the query, therefore, we do not get the predicates as outputs, but values that satisfy them, so that the query is true.}

\vspace{-1mm}
\section{Final Remarks}\label{sec:last}

\vspace{-2mm}
 Explainable data management and explainable AI  (XAI) are effervescent areas of research.
 The relevance of explanations can only grow, as observed from- and due to the legislation and regulations that are being produced and enforced in relation to explainability, transparency and fairness of data management and AI/ML systems.

  Still fundamental research is needed in relation to the notions of {\em explanation} and {\em interpretation}. An always present question is: {\em What is a good explanation?}. \ This is not a new question, and in AI (and other disciplines) it has been investigated. In particular in AI,  areas such as {\em diagnosis} and  {\em causality} have much to contribute.
\ In relation to {\em explanations scores}, there is still a question to be answered: \ {\em What are the desired properties of an explanation score?}
\ignore{The question makes a lot of sense, and may not be beyond an answer. After all, the  general
Shapley value emerged from a list of {\em desiderata} in relation to coalition games, as the only measure that satisfies certain explicit properties \cite{S53,R88}. Although the Shapley value is being used in XAI, in particular in its \shap \ incarnation, there could be a different and specific  set of desired properties of explanation scores that could lead to a still undiscovered explanation score. }

Our work is about interacting with classifiers via answer-set programs. For our work it is crucial to be able to use an implementation of the ASP semantics. We have used {\em DLV}, with which we are more familiar. In principle, we could have used {\em Clingo} instead \cite{clingo}. Those classifiers can be specified directly as a part of the program, as we did in our running example, or they can be invoked by a program as a external predicate \cite{tplp}. From this point of view, our work {\em is not} about learning programs.

We have used in this paper a responsibility score that has a direct origin in {\em actual causality and responsibility}. When the features have many possible values, it makes sense to consider the proportions of value changes that lead to counterfactual versions of the entity at hand, and that of those that do not change the label. In this case, the responsibility score can be generalized to become an average or expected value of label differences \cite{deem,tplp}.

There are different approaches and methodologies in relation to explanations, with causality, counterfactuals and scores being prominent approaches that have a relevant role to play.
Much research is still needed on the use of {\em contextual, semantic and domain knowledge}. Some approaches may be more appropriate in this direction, and we argue that declarative, logic-based specifications can be successfully exploited \cite{tplp}. We have seen how easy becomes adding new knowledge, which would become complicated change of code under procedural approaches.

In this work we have used answer-set programming, in which we have accommodated probabilities as arguments of predicates. Probability computation is done through basic arithmetics provided by the {\em DLV} system. However, it would be more natural to explore the application of probabilistic extensions of logic programming \cite{lucML,riguzzi,lucSRL,kimmig} and   of ASP \cite{pASP}, while retaining the capability to do counterfactual analysis. In this regard, one has to take into account that the complexity of computing the $\xresp$ score is matched by the expressive and computational power of ASP \cite{tplp}.

 \vspace{1mm}{\small  \noindent {\bf Acknowledgments: } \  Part of this work was funded by ANID - Millennium Science Initiative Program - Code ICN17002. Help from  Jessica Zangari and  Mario Alviano with information about  {\em DLV} is very much appreciated.}

\newpage
\appendix

\section{Basics of Answer-Set Programming}\label{sec:asp}

We will give now a brief review of the basics of {\em answer-set programs} (ASPs). As customary, when we talk about ASPs, we refer to {\em disjunctive Datalog programs with weak negation and stable model semantics} \cite{GL91,gelfond}.  For this reason we will, for a given program, use the terms ``stable model" (or simply, ``model") and ``answer-set" interchangeably. \ An answer-set program $\Pi$ consists of a finite number of rules of the form

\vspace{1mm}
\begin{equation}
A_1 \vee \ldots \vee A_n \leftarrow P_1, \ldots, P_m, \nit{not} \ N_1, \ldots, \nit{not} \ N_k, \label{eq:rule}
\end{equation}

\vspace{1mm}\noindent
 where $0\leq n,m,k$, and $A_i, P_j, N_s$ are (positive) atoms, i.e. of the form $Q(\bar{t})$, where $Q$ is a predicate of a fixed arity, say, $\ell$, and $\bar{t}$ is a sequence of length $\ell$ of variables or constants.
In rule (\ref{eq:rule}), $A_1, \ldots, \nit{not} \ N_k$ are called {\em literals}, with $A_1$ {\em positive}, and $\nit{not} \  N_k$, {\em negative}. All the variables in the $A_i, N_s$ appear among those
in the $P_j$.   The left-hand side of a rule is called the {\em head}, and the right-hand side, the {\em body}.  A rule can be seen as a (partial) definition of the predicates in the head (there may be other rules with the same predicates in the head).

The constants in  program $\Pi$ form the (finite) Herbrand universe $H$ of the program. The ground version of
program $\Pi$, $\nit{gr}(\Pi)$, is obtained by instantiating the variables in $\Pi$ in all
possible ways  using
values from $H$. The Herbrand base, $\nit{H\!B}$, of $\Pi$ contains all the atoms obtained as instantiations of
predicates in $\Pi$ with constants in $H$.

A subset $M$ of $\nit{HB}$ is a model of $\Pi$ if it satisfies $\nit{gr}(\Pi)$, i.e.: For every
ground rule $A_1 \vee \ldots \vee A_n$ $\leftarrow$ $P_1, \ldots, P_m,$ $\nit{not} \ N_1, \ldots,
\nit{not} \ N_k$ of $\nit{gr}(\Pi)$, if $\{P_1, \ldots, P_m\}$ $\subseteq$ $M$ and $\{N_1, \ldots, N_k\} \cap M = \emptyset$, then
$\{A_1, \ldots, A_n\} \cap M \neq \emptyset$. $M$ is a minimal model of $\Pi$ if it is a model of $\Pi$, and $\Pi$ has no model
that is properly contained in $M$. $\nit{MM}(\Pi)$ denotes the class of minimal models of $\Pi$.
Now, for $S \subseteq \nit{HB}(\Pi)$, transform $\nit{gr}(\Pi)$ into a new, positive program $\nit{gr}(\Pi)^{\!S}$ (i.e.\  without $\nit{not}$), as follows:
Delete every rule  $A_1 \vee \ldots \vee A_n \leftarrow P_1, \ldots,P_m, \nit{not} \ N_1,$ $ \ldots,
\nit{not} \ N_k$ for which $\{N_1, \ldots, N_k\} \cap S \neq \emptyset$. Next, transform each remaining rule $A_1 \vee \ldots \vee A_n \leftarrow P_1, \ldots, P_m,$ $\nit{not} \ N_1, \ldots,
\nit{not} \ N_k$ into $A_1 \vee \ldots \vee A_n \leftarrow P_1, \ldots, P_m$. Now, $S$ is a {\em stable model} of $\Pi$ if $S \in \nit{MM}(\nit{gr}(\Pi)^{\!S})$.
Every stable model of $\Pi$ is also a minimal model of $\Pi$. Stable models are also commonly called  {\em answer sets}, and so are we going to do most of the time.

 A program is {\em unstratified} if there is a cyclic, recursive definition of a predicate that involves negation. For example, the program consisting of the rules $a \vee b \leftarrow c, \nit{not} \ d$; \ $d \leftarrow e$, and $e \leftarrow b$ is unstratified, because there is a negation in the mutually recursive definitions of $b$ and $e$. The program in Example \ref{ex:hcf} below is not unstratified, i.e. it is {\em stratified}. A good property of stratified programs is that the models can be upwardly computed following {\em strata} (layers) starting from the {\em facts}, that is from the ground instantiations of rules with empty bodies (in which case the arrow is usually omitted). We refer the reader to \cite{gelfond} for more details.

 Query answering under the ASPs comes in two forms. Under the {\em brave semantics}, a query posed to the program obtains as answers those that hold in {\em some} model of the program. However, under the {\em skeptical} (or {\em cautious}) semantics, only the answers that simultaneously hold in {\em all} the models are returned. Both are useful depending on the application at hand.

 \begin{example} \label{ex:hcf} Consider the following  program $\Pi$ that is already ground.

\vspace{-6mm}
\begin{multicols}{2}
\begin{eqnarray*}
a \vee b &\leftarrow& c\\
d &\leftarrow& b\\
a \vee b &\leftarrow& e, \ {\it not} {\it f}\\
e &\leftarrow&
\end{eqnarray*}

\phantom{o}

The program has two stable models: \ $S_1 = \{e, a\}$ and $S_2 = \{e,b,d\}$.

Each of them expresses that the atoms in it are true, and any other atom that does not belong to it, is false.
\end{multicols}

These models are incomparable under set inclusion, and are minimal models in that any proper subset of any of them is not a model of the program (i.e. does not satisfy the program). \boxtheorem
\end{example}

\section{The Complete Example with DLV}\label{sec:example}

{\footnotesize
\begin{verbatim}
 % DLV-COMPLEX
  #include<ListAndSet>
  #maxint = 100000000.
 % domains for program    naiveBayes.txt
    dom_o(sunny). dom_o(overcast). dom_o(rain). dom_t(high). dom_t(medium).
    dom_t(low). dom_h(high). dom_h(normal). dom_w(strong). dom_w(weak).
 % entity schema, naming features, useful to collect info:
    entSchema(outlook,temperature,humidity,wind).
 % original entity that gets label no:
    ent(e,rain,high,normal,weak,o).
 % absolute probabilities for Play (as percentage)
    p(yes, 64). p(no, 36).
 % Outlook conditional probabilities (as percentage)
    p_o_c(sunny, yes, 22). p_o_c(overcast, yes, 45). p_o_c(rain, yes, 33).
    p_o_c(sunny, no, 60). p_o_c(overcast, no, 0). p_o_c(rain, no, 40).
 % Temperature conditional probabilities (as percentage)
    p_t_c(high, yes, 22). p_t_c(medium, yes, 45). p_t_c(low, yes, 33).
    p_t_c(high, no, 40). p_t_c(medium, no, 40). p_t_c(low, no, 20).
 % Humidity conditional probabilities (as percentage)
    p_h_c(normal, yes, 67). p_h_c(high, yes, 33).
    p_h_c(normal, no, 20). p_h_c(high, no, 80).
 % Wind conditional probabilities (as percentage)
    p_w_c(strong, yes, 33). p_w_c(weak, yes, 67).
    p_w_c(strong, no, 60). p_w_c(weak, no, 40).
\end{verbatim}}

{\footnotesize
\begin{verbatim}
 % naive Bayes numerator
   prob_1(E,O,T,H,W,V,Ap) :- ent(E,O,T,H,W,tr), p_o_c(O, V, P1),
                             p_t_c(T, V, P2), A = P1*P2, Ap = A/10,
                             #int(A), #int(Ap), p(V, D).
   prob_2(E,O,T,H,W,V,Bp) :- ent(E,O,T,H,W,tr), prob_1(E,O,T,H,W,V,Ap),
                             p_h_c(H, V, P3), B = Ap*P3, Bp = B/10,
                             #int(B), #int(Bp), p(V, D).
   prob_3(E,O,T,H,W,V,Cp) :- ent(E,O,T,H,W,tr), prob_2(E,O,T,H,W,V,Bp),
                             p_w_c(W, V, P4), C = Bp*P4, Cp = C/10,
                             #int(C), #int(Cp), p(V, D).
   pb_num(E,O,T,H,W,V,Fp) :- ent(E,O,T,H,W,tr), prob_3(E,O,T,H,W,V,Cp),
                             p(V, D), F = Cp*D, Fp = F/10,
                             #int(F), #int(Fp).
                             \end{verbatim}}

{\footnotesize
\begin{verbatim}
 % transition rules: the initial entity or one affected by an intervention
   ent(E,O,T,H,W,tr) :- ent(E,O,T,H,W,o).
   ent(E,O,T,H,W,tr) :- ent(E,O,T,H,W,do).
   \end{verbatim}}

   {\footnotesize
\begin{verbatim}
 % spec of the classifier
   cls(E,O,T,H,W,yes) :- ent(E,O,T,H,W,tr),  pb_num(E,O,T,H,W,yes,Fyes),
                         pb_num(E,O,T,H,W,no,Fno), Fyes >= Fno.
   cls(E,O,T,H,W,no)  :- ent(E,O,T,H,W,tr),  pb_num(E,O,T,H,W,yes,Fyes),
                         pb_num(E,O,T,H,W,no,Fno), Fyes < Fno.
\end{verbatim}}

{\footnotesize
\begin{verbatim}
 % counterfactual rule: alternative single-value changes
   ent(E,Op,T,H,W,do) v ent(E,O,Tp,H,W,do) v
   ent(E,O,T,Hp,W,do) v ent(E,O,T,H,Wp,do) :-
                                          O != Op, T != Tp, H!= Hp, W!= Wp,
                                     ent(E,O,T,H,W,tr), cls(E,O,T,H,W,yes),
                                chosen_o(O,T,H,W,Op), chosen_t(O,T,H,W,Tp),
                                chosen_h(O,T,H,W,Hp), chosen_w(O,T,H,W,Wp),
                                dom_o(Op), dom_t(Tp), dom_h(Hp), dom_w(Wp).
 % definitions of chosen operators:
   chosen_o(O,T,H,W,U) :- ent(E,O,T,H,W,tr), cls(E,O,T,H,W,yes), dom_o(U),
                          U != O, not diffchoice_o(O,T,H,W,U).
   diffchoice_o(O,T,H,W,U) :- chosen_o(O,T,H,W, Up), U != Up, dom_o(U).
   chosen_t(O,T,H,W,U) :- ent(E,O,T,H,W,tr), cls(E,O,T,H,W,yes), dom_t(U),
                          U != T, not diffchoice_t(O,T,H,W,U).
   diffchoice_t(O,T,H,W,U) :- chosen_t(O,T,H,W, Up), U != Up, dom_t(U).
   chosen_h(O,T,H,W,U) :- ent(E,O,T,H,W,tr), cls(E,O,T,H,W,yes), dom_h(U),
                          U != H, not diffchoice_h(O,T,H,W,U).
   diffchoice_h(O,T,H,W,U) :- chosen_h(O,T,H,W, Up), U != Up, dom_h(U).
   chosen_w(O,T,H,W,U) :- ent(E,O,T,H,W,tr), cls(E,O,T,H,W,yes), dom_w(U),
                          U != W,  not diffchoice_h(O,T,H,W,U).
   diffchoice_w(O,T,H,W,U) :- chosen_h(O,T,H,W, Up), U != Up, dom_w(U).

 % not going back to initial entity
   :- ent(E,O,T,H,W,do), ent(E,O,T,H,W,o).

 % stop when label has been changed:
   ent(E,O,T,H,W,s) :- ent(E,O,T,H,W,do), cls(E,O,T,H,W,no).

 % extra denial for not showing models where label does not change
    :- ent(E,O,T,H,W,o), not entAux(E).
 % needs auxiliary predicate, to avoid unsafe negation
   entAux(E) :- ent(E,O,T,H,W,s).
\end{verbatim}}
 {\footnotesize
\begin{verbatim}
 % collecting changed values for each feature:
   expl(E,outlook,O)  :- ent(E,O,T,H,W,o), ent(E,Op,Tp,Hp,Wp,s), O != Op.
   expl(E,temp,T)     :- ent(E,O,T,H,W,o), ent(E,Op,Tp,Hp,Wp,s), T != Tp.
   expl(E,humidity,H) :- ent(E,O,T,H,W,o), ent(E,Op,Tp,Hp,Wp,s), H != Hp.
   expl(E,wind,W)     :- ent(E,O,T,H,W,o), ent(E,Op,Tp,Hp,Wp,s), W != Wp.
\end{verbatim}}

{\footnotesize
\begin{verbatim}
 % forming and collecting contingency sets
   cause(E,U)       :- expl(E,U,X).
   cauCont (E,U,I)  :- expl(E,U,X), expl(E,I,Z), U != I.
   preCont(E,U,{I}) :- cauCont(E,U,I).
   preCont(E,U,#union(Co,{I})) :- cauCont(E,U,I), preCont(E,U,Co),
                                  not #member(I,Co).
   cont(E,U,Co)     :- preCont(E,U,Co), not HoleIn(E,U,Co).
   HoleIn(E,U,Co)   :- preCont(E,U,Co), cauCont(E,U,I),not #member(I,Co).
   tmpCont(E,U)     :- cont(E,U,Co), not #card(Co,0).
   cont(E,U,{})     :- cause(E,U), not tmpCont(U).

 % computing the inverse of Resp
   invResp(E,U,R) :- cont(E,U,S), #card(S,M), R = M+1, #int(R).
 % full explanations:
   fullExpl(E,U,R,S) :-  expl(E,U,X), cont(E,U,S), invResp(E,U,R).
\end{verbatim}}

{\footnotesize
\begin{verbatim}
 Synthesized output

 M1 {ent(e,rain,high,normal,weak,o), ent(e,rain,high,normal,weak,tr),...,
    pb_num(e,rain,high,normal,weak,yes,20665),
    pb_num(e,rain,high,normal,weak,no,4608),
    cls(e,rain,high,normal,weak,yes), ..., ent(e,rain,high,high,weak,do),
    ent(e,rain,high,high,weak,tr), ...,
    pb_num(e,rain,high,high,weak,yes,10156),
    pb_num(e,rain,high,high,weak,no,18432),
    cls(e,rain,high,high,weak,no), ent(e,rain,high,high,weak,s), ...,
    expl(e,humidity,normal), cont(e,humidity,{}), invResp(e,humidity,1),
    fullExpl(e,humidity,1,{})}

 M2 {ent(e,rain,high,normal,weak,o), ent(e,rain,high,normal,weak,tr),
    chosen_h(rain,high,normal,weak,high),
    chosen_w(rain,high,normal,weak,strong),
    ent(e,rain,high,high,strong,do), ent(e,rain,high,high,strong,tr),
    pb_num(e,rain,high,high,strong,yes,5004),
    pb_num(e,rain,high,high,strong,no,27648),
    cls(rain,high,high,strong,no), ent(e,rain,high,high,strong,s),
    expl(e,humidity,normal), expl(e,wind,weak), cauCont(e,humidity,wind),
    cauCont(e,wind,humidity),cause(e,humidity), cause(e,wind), entAux(e),
    preCont(e,humidity,{wind}), preCont(e,wind,{humidity}),
    cont(e,humidity,{wind}),cont(e,wind,{humidity}), tmpCont(e,humidity),
    tmpCont(e,wind), invResp(e,humidity,2), invResp(e,wind,2),
    fullExpl(e,humidity,2,{wind}), fullExpl(e,wind,2,{humidity})}

 M3 {ent(e,rain,high,normal,weak,o), ent(e,sunny,high,normal,strong,tr),
    pb_num(e,sunny,high,normal,strong,yes,6777),
    pb_num(e,sunny,high,normal,strong,no,10368),
    cls(e,sunny,high,normal,strong,no), ent(e,sunny,high,normal,strong,s),
    invResp(e,outlook,2), fullExpl(e,outlook,2,{wind}), ...}

 M4 {ent(e,rain,high,normal,weak,o), ent(e,sunny,high,high,weak,tr),
    pb_num(e,sunny,high,high,weak,yes,6771),
    pb_num(e,sunny,high,high,weak,no,27648),
    cls(e,sunny,high,high,weak,no), ent(e,sunny,high,high,weak,s),
    invResp(e,outlook,2), fullExpl(e,outlook,2,{humidity}), ...}

 M5 {ent(e,rain,high,normal,weak,o), ent(e,rain,medium,high,strong,tr),
    pb_num(e,rain,medium,high,strong,yes,10304),
    pb_num(e,rain,medium,high,strong,no,27648),
    cls(e,rain,medium,high,strong,no), ent(e,rain,medium,high,strong,s),
    invResp(e,temp,3), fullExpl(e,temp,3,{humidity,wind}), ...}

M6 {ent(e,rain,high,normal,weak,o), ent(e,rain,low,high,strong,tr),
    pb_num(e,rain,low,high,strong,yes,7513),
    pb_num(e,rain,low,high,strong,no,13824),
    cls(e,rain,low,high,strong,no), ent(e,rain,low,high,strong,s),
    invResp(e,temp,3), fullExpl(e,temp,3,{humidity,wind}), ...}

 M7 {ent(e,rain,high,normal,weak,o), ent(e,sunny,low,high,weak,tr),
    pb_num(e,sunny,low,high,weak,yes,10156),
    pb_num(e,sunny,low,high,weak,no,13824),
    cls(e,sunny,low,high,weak,no), ent(e,sunny,low,high,weak,s),
    invResp(e,outlook,3),fullExpl(e,outlook,3,{humidity,temp}), ...}

 M8 {ent(e,rain,high,normal,weak,o), ent(e,sunny,medium,high,weak,tr),
    pb_num(e,sunny,medium,high,weak,yes,13977),
    pb_num(e,sunny,medium,high,weak,no,27648),
    cls(e,sunny,medium,high,weak,no), ent(e,sunny,medium,high,weak,s),
    invResp(e,outlook,3), fullExpl(e,outlook,3,{humidity,temp}), ...}

 M9 {ent(e,rain,high,normal,weak,o), ent(e,sunny,medium,high,strong,tr),
    pb_num(e,sunny,medium,high,strong,yes,6880),
    pb_num(e,sunny,medium,high,strong,no,41472),
    cls(e,sunny,medium,high,strong,no), ent(e,sunny,medium,high,strong,s),
    invResp(e,outlook,4), fullExpl(e,outlook,4,{humidity,temp,wind}), ...}

 M10 {ent(e,rain,high,normal,weak,o), ent(e,sunny,low,high,strong,tr),
     pb_num(e,sunny,low,high,strong,yes,5004),
     pb_num(e,sunny,low,high,strong,no,20736),
     cls(e,sunny,low,high,strong,no), ent(e,sunny,low,high,strong,s),
     invResp(e,outlook,4),fullExpl(e,outlook,4,{humidity,temp,wind}), ...}
\end{verbatim} }

{\footnotesize
\begin{verbatim}
 Queries

 % Uncomment one at a time, or only the last one will be answered
  % Q1 responsibility for non-maximum feature outlook, brave
   % invResp(e,outlook,R)?

  % Q2 explanations requiring less than 3 changes, brave
   % fullExpl(E,U,R,S), R<3?

  % Q3 combination exists, brave
   % cls(E,O,T,H,W,_), O=sunny, W=strong?

  % Q4 entities with switched label, brave
   % cls(E,O,T,H,W,no)?

  % Q5 combinations that don't change feature Wind, cautious and brave
   % ent(e,_,_,_,Wp,s), ent(e,_,_,_,W,o), W = Wp?
\end{verbatim} }

{\footnotesize
\begin{verbatim}
 Query outputs

  % Q1 responsibility for non-maximum feature outlook, brave
     2
     3
     4

  % Q2 explanations requiring less than 3 changes, brave
     e, outlook, 2, {humidity}
     e, outlook, 2, {wind}
     e, humidity, 1, {}
     e, humidity, 2, {wind}
     e, humidity, 2, {outlook}
     e, wind, 2, {humidity}
     e, wind, 2, {outlook}

  % Q3 combination exists, brave
     e, sunny, high, normal, strong, no
     e, sunny, medium, high, strong, no
     e, sunny, medium, normal, strong, yes
     e, sunny, low, high, strong, no
     e, sunny, low, normal, strong, yes


  % Q4 entities with switched label, brave
     e, rain, high, high, weak
     e, sunny, low, high, strong
     e, sunny, medium, high, strong
     e, rain, low, high, strong
     e, rain, medium, high, strong
     e, sunny, high, normal, strong
     e, rain, high, high, strong
     e, sunny, medium, high, weak
     e, sunny, low, high, weak
     e, sunny, high, high, weak

  % Q5 combinations that don't change feature Wind, cautious (empty)
     DLV [build BEN/Jul 13 2011   gcc 4.5.2]

  % Q5 combinations that don't change feature Wind, brave
     rain, high, high, weak, rain, high, normal, weak
     sunny, medium, high, weak, rain, high, normal, weak
     sunny, high, high, weak, rain, high, normal, weak
     sunny, low, high, weak, rain, high, normal, weak
\end{verbatim} }


We can add weak constraints to minimize the number of changes of feature values:
{\footnotesize
\begin{verbatim}
 % weak constraints to minimize number of changes
   :~ ent(E,O,T,H,W,o), ent(E,Op,Tp,Hp,Wp,s), O != Op.
   :~ ent(E,O,T,H,W,o), ent(E,Op,Tp,Hp,Wp,s), T != Tp.
   :~ ent(E,O,T,H,W,o), ent(E,Op,Tp,Hp,Wp,s), H != Hp.
   :~ ent(E,O,T,H,W,o), ent(E,Op,Tp,Hp,Wp,s), W != Wp.
\end{verbatim}}

Running the program with them, leaves only the model \verb+M1+ above, corresponding to the counterfactual entity $\e' = \nit{ent}(\mathsf{rain},\mathsf{high},\mathsf{high},\mathsf{weak})$. This is a maximum-responsibility counterfactual explanation.

\end{document}